# VOXEL-BASED POINT CLOUD LOCALIZATION FOR SMART SPACES MANAGEMENT

F. S. Mortazavi *, O. Shkedova, U. Feuerhake, C. Brenner, M. Sester

Institute of Cartography and Geoinformatics, Leibniz University, Hannover, Germany- (Faezeh.Mortazavi, Olga.Shkedova, Udo.Feuerhake, Claus.Brenner, Monika.Sester)@ikg.uni-hannover.de

**ABSTRACT:**

This paper proposes a voxel-based approach for creating a digital twin of an urban environment that is capable of efficiently managing smart spaces. The paper explains the registration and localization procedure of the point cloud dataset, which uses the KISS ICP for scan point cloud combination and the RANSAC method for the initial alignment of the combined point cloud. The mobile mapping point cloud using Riegl VMX-250 serves as the reference map, and Velodyne scans are used for localization purposes. The point-to-plane iterative closest-point method is then employed to refine the alignment. The paper evaluates the efficacy of the proposed method by calculating the errors between the estimated and ground truth positions. The results indicate that the voxel-based approach is capable of accurately estimating the position of the sensor platform, which are applicable for various use cases. A specific use case in the context is smart parking space management, which is described and initial visualization results are shown.

## 1. INTRODUCTION

The last decades' improvement of instruments and methods for three-dimensional (3D) geodata collection, analysis, and representation gives an opportunity for an extensive range of applications. The 3D geoinformation has become a basis for crisis and disaster management, environmental simulations, facility management, and urban planning (Saranet al., 2018). Additionally, it plays a significant role in the operation processes of modern smart city concepts. Most data related to the city topography, communication systems, spatial arrangement, infrastructure organization, city resources, and points of interest distribution directly or indirectly require geographical referencing for efficient functioning (Gotlib and Olszewski, 2021). The main components of a smart city are defined as mobility, governance, environment, and people that ensure the functionality of demanded services such as healthcare, transportation, education, and energy, the amount of which grows permanently (Al Nuaimi et al., 2015). Therefore, the promise of today's smart digital city leads to a significant increase in data quantity and systems complexity (Hashem et al., 2016). Considering the aforementioned requirements, an urban digital twin was introduced as a smart city concept innovation for the development of integrated and intelligent systems. In that frame, the usage of diverse data from numerous sensors, the design of an adaptive digital model that learns from and evolves with the real city, and the elaboration of predictive models are realized for future scenarios anticipation capability. (Castelli et al., 2022). The idea of digital twins first appeared in the context of lifecycle management product modelling at the University of Michigan in 2002 (Grieves and Vickers, 2016). It was represented as mirrored real space in a digital or virtual space with data and information exchange between these two environments. However, the term digital twin was defined later, in 2010, in a NASA integrated technology roadmap related to flying vehicles, as "an integrated multi-physics, multi-scale, probabilistic simulation of a vehicle or system that uses the best available physical models, sensor updates, fleet history, etc., to mirror the life of its flying twin" (Shafto et al., 2010). The urban digital twin takes the begging from 3D city construction models, which were advanced by joining existing digital representations of buildings and infrastructure within Building Information Modelling (BIM). The continuous, bidirectional data transmission between real and virtual space property gives an opportunity to model a city in the most accurate form (Ferré-Bigorra et al., 2022). Moreover, the 3D geodata exchange allows for spatial relationship description and environment updates according to the real world at a given time (Bacher, 2022). Beside BIM and landscape planning, the CityGML framework was released for the representation, storage, and arrangement of virtual 3D city models (Scalas et al., 2022).

Recently, a number of urban digital twins were implemented. The digital twin of Zurich (Switzerland) includes a 3D city model that consists of blocks, roofs, terrain, street space, trees, archaeological objects, power lines, and bridges. (Schrotter and Hürzeler, 2020). The National Research Foundation established an urban digital twin for Singapore city. Virtual Singapore is a dynamic 3D city model that comprises detailed information, such as material, the texture of geometrical objects, terrain attributes, and models of building compounded from walls, floors, and ceilings. (Virtual Singapore, 2018). Furthermore, to improve the realism of the Singapore gardens visualization, the CityGML tree models framework with multiple levels of details (LoD) was elaborated (Gobeawan et al., 2018). The Hong Kong Science and Technology Parks Corporation, together with Chain Technology Development, has provided the digital twin solution called the Hybrid Reality Platform for Hong Kong Science Park. The aim of the project is to install a data management system, which is capable of large-sized BIM objects and 3D reality meshes processing (Yang, Seungho and Kim, 2021). One more 3D city model was generated for Herrenberg (Germany). The urban digital twin of the town was built based on a hybrid of solid 3D models, geographical data, a digital elevation model (DEM), 3D laser scan data, and BIM (Dembski et al., 2020). As can be noticed, the existing urban digital twins are realized within uniform means of solid 3D models, BIM, and CityGML format for the virtual city representation. Despite the high accuracy, frequent updates of such models according to continuous changes in the real world remain challenging. In addition, the current LoD concept of CityGML is not sufficiently flexible for indoor applications (Tang et al., 2018). The correspondence of the urban indoor and outdoor virtual environment to the real-time condition is crucial for emergency service operations, autonomous vehicle route planning, and dynamic smart space management.

---

* Corresponding author

Therefore, this paper proposes an approach different from the standard digital twin formation and management methods. The central idea is capturing the three dimensions of the environment in high resolution using a many of different sensor systems and breaking it down into billions of voxels, systematically addressing and updating them utilizing identification numbers. Thus, the system will be able to combine, modern digital twins' functionality, fast, up-to-date data performance, and open new opportunities for applications. Such a virtual voxel-based representation of the real world will enable people and machines to meet the growing competition in space use in a more demand-oriented, efficient, secure, and fair way, both for themselves and in the interests of the public. In order to fuse data from different sensors, it is necessary to align these data to the current 3D voxel model. To this end efficient localization methods are needed.

## 1.1 Voxel representation

Voxels can be considered as the 3D equivalent of pixels in a two-dimensional (2D) space. Similar to the 2D case, voxels are placed on a 3D grid with uniform spacing in all dimensions. Analogous to a 2D square pixel representation, a voxel corresponds to a 3D cube (Chajdas, 2015).

Voxel representation is used for a multitude of tasks including finite-element simulation, object detection, classification, 3D reconstruction, localization, trajectory planning, computer graphics rendering techniques etc. (Koketsu et al., 2023; Pantaleoni, 2011; Agus et al., 2010; Ma et al. 2021, Mao et al., 2021; Xie et al. 2018). A wide range of voxels utilization can be explained by the properties of this type of representation, such as uniform resolution, regular structure of independent cells that removes the complexities of various computations and manipulations, trivial simplification, and easy data stream (Chajdas, 2015). Voxel-based models are actively applied to define the unexplored space for aerial and terrestrial autonomous vehicles (Oleynikova et al., 2017). Furthermore, voxels are utilized within various data structures for efficient storage and model updates. For instance, Wu et al. (2022) proposed a multi-level voxel representation for a digital twin model of the geological environment, which allows for dynamic updates of complex geological structures. Consequently, the voxel model can be set as a proper basis for multifunctional, complicated systems, and opens promising perspectives for urban digital twin realization.

## 1.2 Localization methods

In the last decades, many methods have been proposed for point cloud registration and localization. Iterative closest point (ICP) is one the most important methods in registration field, which works with iterating the search of corresponding points to minimize the difference between two points clouds (Besl and McKay, 1992). Another famous method is the normal distribution transform (NDT), which assigns a normal distribution to each 2D cell (Biber and Straßer, 2003). This kind of methods can be used for fine registration, where an initial pose estimation is needed. On the other hand, there are some other methods that work without any need of initial alignments, like the random sample consensus (RANSAC) (Schnabel et al., 2007) or the 3D Hough transform (Hulik et al., 2014).

In addition, there are some other methods for global localization to find the position of current point clouds in the reference map, such as global feature matching (Luo et al., 2021), and map matching (Feng et al., 2017).

KISS ICP (Vizzo et al., 2023) is an approach for LiDAR odometry that can accurately compute a robot's pose during navigation. The approach is simple but effective, and its core components include motion prediction and scan de-skewing, spatial scan sub-sampling, an adaptive threshold for correspondence search, and ICP with a robust kernel. In this approach, the point-to-point ICP has been used, and it is able to be comparable to state-of-the-art odometry systems and can accurately compute a robot's odometry in various environments without relying on IMUs or wheel odometers.

In some cases when the initial pose is not available, one of the best solutions for point cloud registration is to use coarse to fine registration, which we will discuss in this paper. In this case, RANSAC is used to estimate the initial pose of scan points for coarse registration, and then the alignment result of RANSAC is used as initialization of fine registration method like ICP.

In this paper, we describe a preliminary voxel-based urban environment representation, generated from point cloud datasets derived from different sensors. In addition, we demonstrate its future applicability for a real-time intelligent parking management use case. On this account, the most attention in the article is given to the registration and localization procedure of point clouds measured by distinct sensors. The paper is organized as follows: first the methodology is described, as well as the data set used. Subsequently, the experiments are conducted and the results are evaluated. Based on the aligned data sets, the application scenario of smart parking is described.

## 2. METHODOLOGY

The objective of this section is to present an approach for localizing Velodyne's raw scans on a map, which is crucial for accurate positioning and updating the map. As mentioned in the introduction, the ICP method is a traditional technique for scan localization. However, it is limited by its reliance on an initial value for alignment. Unfortunately, accessing the initial value is not always feasible. On the other hand, lacking enough features in sparse point clouds is another challenge in localization term. Therefore, a three-step method is proposed in this research work to overcome this limitation. This approach does not rely on an initial value and offers a more robust localization solution. The three consecutive steps of the localization process in this article are:

1) applying voxelization
2) applying RANSAC to coarse registration, using the Fast Point Feature Histogram, based on a coarse grid resolution
3) applying a Pont-to-Plane ICP to fine registration, based on the original voxel size

As shown in Figure 1, in the first step, the raw Velodyne scans are initially processed and combined using the KISS ICP algorithm (Vizzo et al., 2023) to create a denser point cloud of the area of interest. The resulting point cloud is then used to determine an initial alignment value for the Iterative Closest Point (ICP) algorithm. To this end, a robust RANSAC method is applied to localize the combined point cloud on the reference map (MMS). Once the RANSAC process is completed, the resulting transformation matrix is used as the initial alignment for each individual scan in the ICP method. In the subsequent sections, we describe the details of each of these steps.

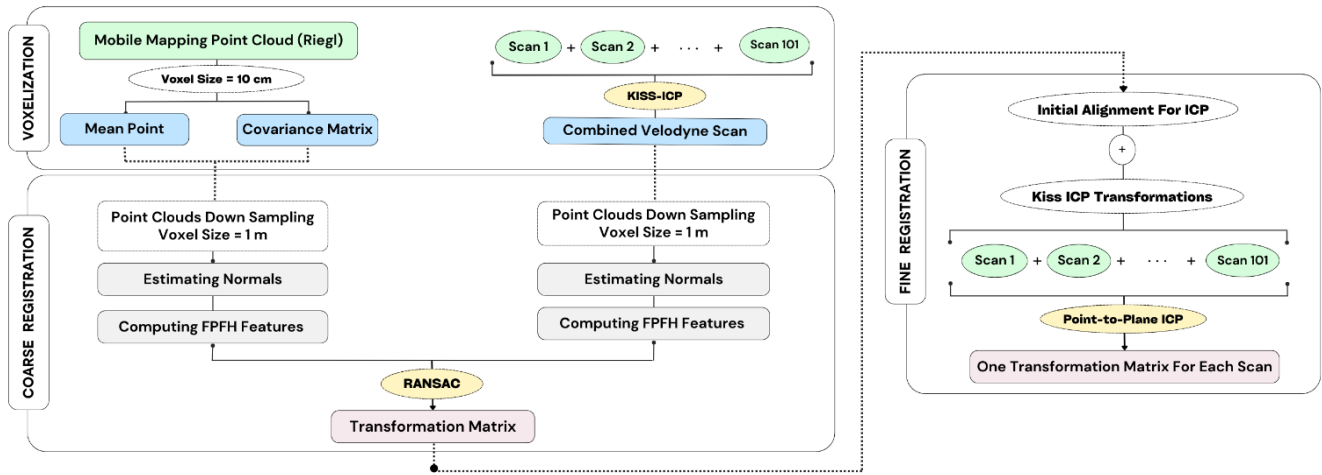

**Figure 1**. The pipeline of localization process

## 2.1 Datasets

This study focuses on analyzing a point cloud dataset collected along a residential street (Am Kleinen Felde) in the city of Hanover, Germany, covering 175 meters.

The dataset consists of two separate parts: a mobile mapping point cloud provided by a Riegl VMX-250 system (MMS) (Figure 2.top), which is used as reference map, and scans from a Velodyne VLP 16 (Velodyne) (Figure 2.bottom), which are to be localized.

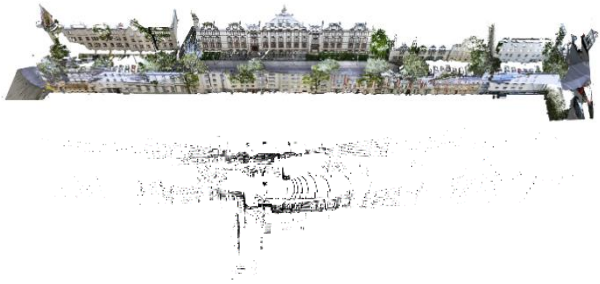

**Figure 2**. MMS point cloud with colors (top), a Velodyne scan from top view (bottom)

The Riegl VMX-250 technology is used to collect a dense point cloud consisting of approx. 4,340,000 points. It is an advanced laser scanning system equipped with four cameras that provide color information, allowing to colorize the point cloud. The Velodyne scans (101 scans) are used to update the map, where each scan containing around 25,000 points. The point clouds have been measured at different times, the mobile mapping data has been obtained earlier than the Velodyne scans. Thus, the goal is to localize the Velodyne scans in the reference map generated from the Mobile Mapping point cloud.

## 2.2 Voxelization

During the voxelization process, a regular 3D grid with a given spacing is created from the acquired reference dense point cloud (MMS); in this case, a spacing of 10cm was selected. Therefore, a cube with a 10 cm edge length describes each voxel. Every voxel of the 3D grid is assigned a unique index number and contains the corresponding points of the reference point cloud. Next, the necessary information for the further processing is obtained.

Firstly, for the localization and registration step, the mean point per voxel is computed from all the reference points inside. Considering the 3D coordinate system, the mean point coordinates are calculated. Afterwards, the covariance matrices $C$ are calculated according to the covariance values $\sigma$ between all the dimensions:

$$\sigma(x, y) = \frac{1}{n-1}\sum_{i=1}^{n}(x_i - \overline{x})(y_i - \overline{y}) \tag{1}$$

$$C = \begin{bmatrix} \sigma(x, x) & \sigma(x, y) & \sigma(x, z) \\ \sigma(y, x) & \sigma(y, y) & \sigma(y, z) \\ \sigma(z, x) & \sigma(z, y) & \sigma(z, z) \end{bmatrix} \tag{2}$$

where, $x_i, y_i, z_i$ = coordinates of reference point
$n$ = total number of reference points inside the voxel
$\overline{x}, \overline{y}, \overline{z}$ = coordinates of the mean point inside the voxel

The derived covariance matrices allow to determine the surface normal, a linear, planar and other features estimation. Additionally, the mean RGB color value for every voxel is defined.

Lacking enough features to identify objects make object recognition difficult. To overcome this challenge, we propose using a batch of scans as a more effective alternative. By utilizing the KISS ICP method, we can create a consolidated block of scans that can be used to accurately locate objects on the map (Figure 3). Using information from multiple scans allows us to have a more comprehensive understanding of the environment.

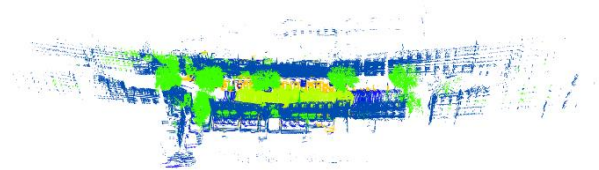

**Figure 3**. Combined Velodyne point cloud using KISS ICP.

Once the KISS ICP algorithm is performed, a transformation matrix is generated for each scan, representing the spatial transformation of that scan with respect to the first scan, which serves as the starting point (0, 0, 0). These transformation matrices can be used to access the combined Velodyne scan.

## 2.3 Coarse registration

Determining the initial value has been always a significant challenge in the localization field, and several methods have been proposed to overcome this problem. One of the commonly used methods for rough localization is the Random Sample Consensus (RANSAC) algorithm. However, applying RANSAC to each Velodyne scan separately can be time-consuming, and there are limited features in Velodyne scans alone.

Since high accuracy results is not required in the coarse registration section, to reduce the computation time, the first step is to down sample the input data, which includes the mobile mapping data and the combined Velodyne scan to 1-meter voxels. Using 1-meter voxels for coarse registration will be faster than using 10 cm voxels due to the reduced number of voxels. Furthermore, the registration results achieved with 1-meter voxels are sufficient to be used as an initial transformation matrix for fine registration.This voxelization process significantly reduces the amount of data and simplifies subsequent feature extraction. The calculation of the normal vector for each voxel is then performed using the covariance matrix. The normal vectors are crucial for calculating the Fast Point Feature Histogram (FPFH) for each voxel.

The FPFH is a feature descriptor that characterizes the local geometric properties of a point in a three-dimensional space (Rusu et al., 2009). This approach involves the application of a pre-defined radius to identify the nearest neighbors of each data point, within a three-dimensional spherical region of radius r. In the next step, the relationships between each query point and its neighbors are defined. For each pair of points ($p_i$ and $p_j$) and their normal ($n_i$ and $n_j$) the following angular variations are computed which is called Simplified Point Feature Histogram (SPFH):

$$\alpha = v.n_j$$
$$\varphi = \left(u.\left(p_j - p_i\right)\right)/\|p_j - p_i\|$$
$$\theta = \arctan\left(w.n_j, u.n_j\right) \tag{3}$$

Where $u = n_i$, $v = \left(p_j - p_i\right) \times u$ and $w = u \times v$.

In the following step, a neighborhood search is performed for every neighboring point by recalculating its proximity criteria based on the previous radius. Once the neighboring points are identified, their SPFH descriptors are computed to characterize their geometrical properties, which provide a robust representation of the local surface structure around each point. Subsequently, a weighted histogram is constructed by combining the SPFH descriptors of each neighboring point to create a holistic description of the local surface geometry. As shown in Figure 4, the red query point is linked only to its neighbors (k-points), which are enclosed by the gray circle.

The direct neighbors are then linked to their respective neighbors, and the histograms resulting from the direct neighbors and the histogram of the query point are combined with weights to create the FPFH.

$$FPFH(p) = SPF(p) + \frac{1}{k}\sum_{i=1}^{k}\frac{1}{w_k}.SPF(p_k) \tag{4}$$

The value of $w_k$ indicates the distance between the reference point $p$ and a neighboring point $p_k$.

A global registration is achieved by utilizing the RANSAC algorithm, whereby a set of points is randomly selected from the combined Velodyne scan during each iteration. The corresponding points in the MMS point cloud are identified by searching for the nearest neighbor in the FPFH feature space. To efficiently evaluate a large number of potential correspondences, numerous candidate correspondences are sampled and ranked rapidly based on the similarity between their corresponding histograms.

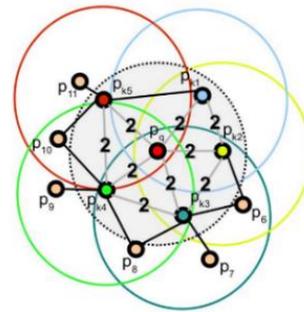

**Figure 4**. The relationships between query point and its neighbors (Rusu et al., 2009).

## 2.4 Fine registration

In order to reduce the time of computations, the global registration is limited to 1-meter down-sampled point cloud, which could potentially lead to a less precise alignment. While the initial alignment obtained through the RANSAC method may not be accurate enough, it can serve as a good initial transformation value for the Point-to-plane Iterative Closest Point (ICP) algorithm. The Point-to-plane ICP is a variation of the traditional ICP that takes a different approach to distance calculations. Rather than computing the distance between point pairs, the algorithm measures the distance between a point in one point cloud and the plane tangent to the surface of the other point cloud at the corresponding point. By iteratively repeating this process, a more accurate alignment between the two point clouds can be achieved. During this step, instead of utilizing the combined Velodyne scan, the focus is shifted towards using individual raw Velodyne scans for localization purposes. To accomplish this, the transformation matrices obtained from KISS ICP are applied to the raw point clouds of the Velodyne scans. Following this, the transformation matrix obtained from RANSAC is employed as an initial value for the point-to-plane ICP process on each of the individual scans. At the end of this process, each raw Velodyne scan is successfully localized to our global map (MMS), which can help us to accurately navigate and analyse the mapped environment.

## 3. EXPRIMENTS AND RESULTS

### 3.1 Voxels

The voxels of 10 centimeters in size are generated to form the regular 3D grid. As described in the methodology, for each voxel the mean point, RGB color value and covariance matrix are calculated. The computed parameters enable the different feature estimation and the real-life, "true" color visualization. Figure 5 shows the voxel-based urban environment within "true" colors.

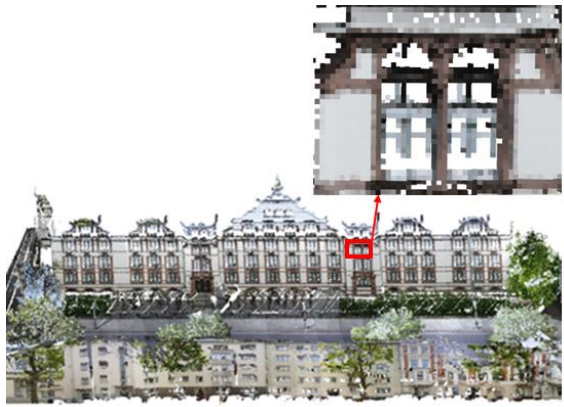

**Figure 5**. Voxel-based urban environment

### 3.2 Localization

As explained in the methodology section, the KISS ICP method is used for merging multiple point cloud scans. The outcome of this method is a dense and unified point cloud. In addition to the transformation matrix generated for each individual scan, the unified point cloud can be further processed for point cloud classification. For this purpose, the Kernel Point Convolution (KPConv) network is employed to classify the combined Velodyne scan, as shown in Figure 3. The generated labels can be used in the future for map updating and other applications. The ability to classify points in the unified point cloud facilitates the recognition of objects and features in the environment, which is essential for extracting meaningful information.

As described in the methodology, in the subsequent phase, a hierarchical approach is applied to down-sample the combined Velodyne scan to 1-meter voxels. This approach ensures consistency as the 10 cm voxels in the MMS point cloud is also converted to 1-meter voxels. The resulting down-sampled point cloud speeds up processing and allows for more efficient handling of large volumes of data.

To perform the alignment between the combined point cloud and the reference point cloud, the RANSAC method is used with a distance threshold of 2 meters. However, due to the complexity of the environment, the output of the registration process may not be entirely accurate. As shown in Figure 6, although after coarse registration the combined point cloud is generally aligned with the reference point cloud; there are some residual misalignments in terms of shift and rotation. Despite the slight misalignment, the approximate alignment obtained with RANSAC is still suitable and sufficient as an initial estimate for the iterative closest point (ICP) algorithm to refine the alignment. By using the approximate alignment as an initial estimate, the ICP algorithm converges faster and provides a more accurate final alignment.

In the next step, the point-to-plane ICP method is applied to improve the alignment. This process uses the original dataset, where the down sampled MMS point cloud with 10 cm voxels serve as the reference map, and the single Velodyne scans are used for localization.

As shown in Figure 7, using the transformation matrix from the RANSAC method as the initial value for ICP, the resulting alignment between the raw Velodyne scans and the reference map is quite accurate.

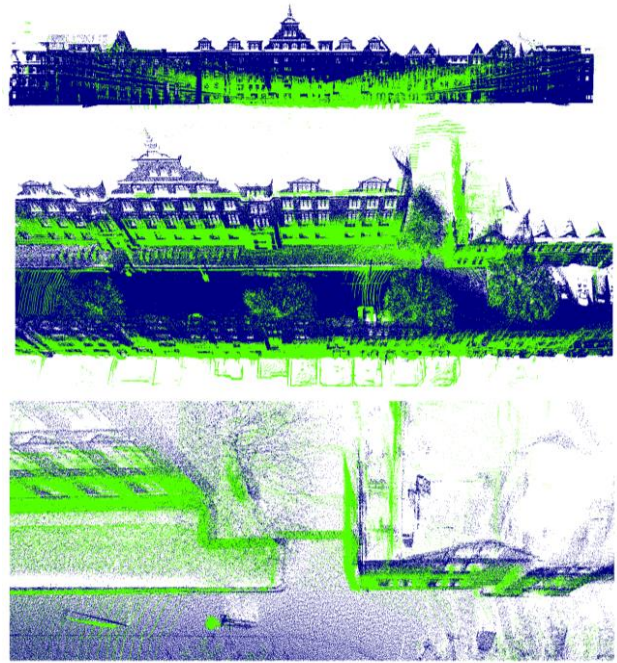

**Figure 6**. Coarse registration result using RANSAC

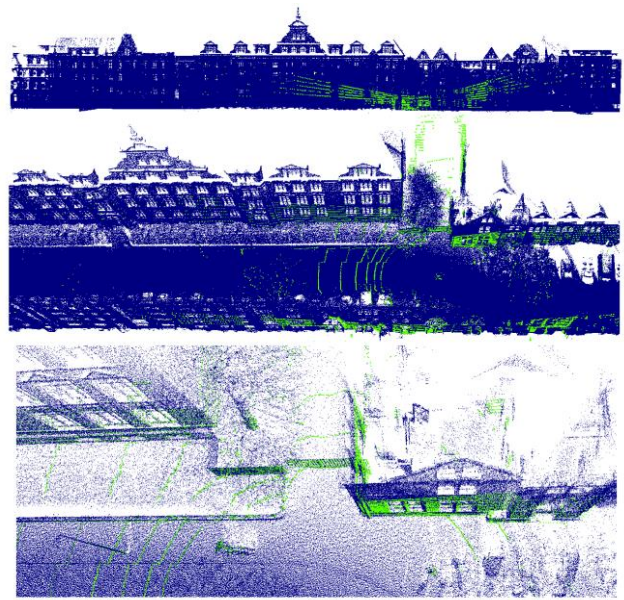

**Figure 7**. Fine registration result using point-to-plane ICP

### 3.3 Evaluation

In this research study, 101 Velodyne scans are utilized for localization. Furthermore, as shown in Figure 8, the estimated positions are represented in red color, and the ground truth values are displayed in green color. The fact that the trajectories of the estimated positions and the ground truth values are very close to each other suggests that the proposed method can accurately estimate the position and orientation of the sensor platform.

To assess the performance and effectiveness of the proposed voxel-based approach, we calculate the errors between the estimated and the ground truth positions for each scan in both the

329

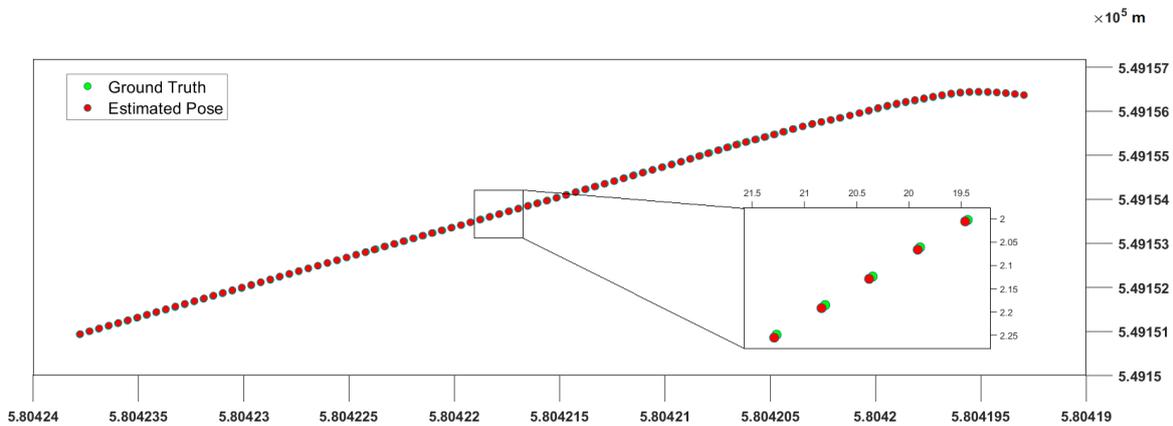

**Figure 8.** Trajectory of estimated pose of 101 scans.

XY and Z directions. Figure 9.a presents the error values as a histogram.

The histogram in the XY direction shows that many of the scans (88 out of 101) have errors less than 3.6 cm. A smaller number of scans (13 scans) have errors greater than this value, but less than 6 cm. The maximum error observed in this direction is 6.3 cm, which occurred in only 3 scans. This suggests that the estimation algorithm performed well overall, with most scans having relatively small errors.

The vast majority of scans in the Z direction have errors of less than 2 cm, with most of them between 0 to 1 cm. Only a small percentage of scans have errors greater than 2 cm, with just four scans having errors between 3 to 5 cm (see Figure 9.b).

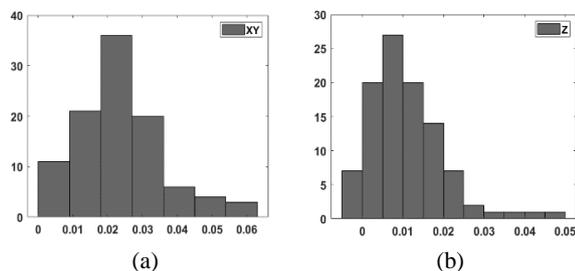

(a)                    (b)

**Figure 9.** The errors between the estimated and ground truth positions for each scan in a) XY and b) Z directions.

## 4. SMART PARKING SPACES MANAGEMENT

Lack of parking spaces, disordered parking, and insufficient parking space use are major problems in big cities nowadays. Therefore, smart parking management is an exemplary use case for urban digital twins. Based on the results of the localization procedure for mobile mapping and Velodyne sensor data measured in different time periods, described in section 3, we have investigated the possibility of parking space allocation and their occupancy examination for a smart parking space management use case within the proposed urban digital twin platform. The idea is to define the initial state of parking spaces according to the dense reference point cloud and to use the sparse, registered Velodyne data for the parking space occupancy state update. To this end, the point clouds of both sensors are classified with respect to major object classes, for example cars, trees, ground etc. Afterwards, the occupied parking spaces are defined by bounding boxes according to the clustered reference point cloud. Finally, the occupancy of the parking spaces is examined by the detection of Velodyne points, classified as cars, inside the determined previously bounding boxes. If the Velodyne points of the class "car" are detected inside the bounding box – the parking space is considered as occupied, if opposite, then the parking space is vacant. A more detailed explanation is provided in the next subsections.

### 4.1 Classification

To effectively distinguish between dynamic and static objects in web applications and visualization, it is essential to have a classified and labelled point cloud. For this purpose, the Kernel Point Convolution (KPConv) is used for point cloud classification (see Figure 10). This method operates directly on point clouds without the need for an intermediate representation. KPConv inputs neighborhoods of a specific radius and processes them using weights that have been located by a small number of kernel points.

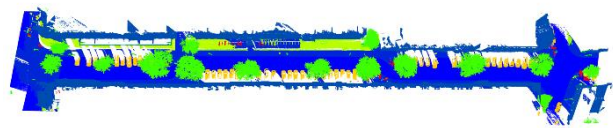

**Figure 10.** The classified MMS point cloud using the Kernel Point Convolution (KPConv) network

### 4.2 Bounding boxes

After classifying the point cloud, the resulting labels can be used to separate different objects from each other. For instance, in parking space management, identifying the points that represent cars is crucial to determine the exact location of each vehicle and whether the parking space is occupied or not.

To extract the cars from the point cloud, the first step is to remove all points that do not represent cars. This can be achieved using the labels obtained from the point cloud classification. Clustering is the next step after extracting the cars from the point cloud. To perform the clustering, the Euclidean distance method is used. This results in each car being represented by a separate cluster of points. However, clustering alone is not enough to accurately identify the shape and position of a car. Therefore, bounding boxes are fitted to each cluster to better represent the actual size and location of the vehicle. This is done by fitting a 3D bounding box around the cluster of points that represent the car (Figure 11). The output data includes the centroid of each box and its dimensions in all three directions (X, Y, and Z), including its length and orientation.

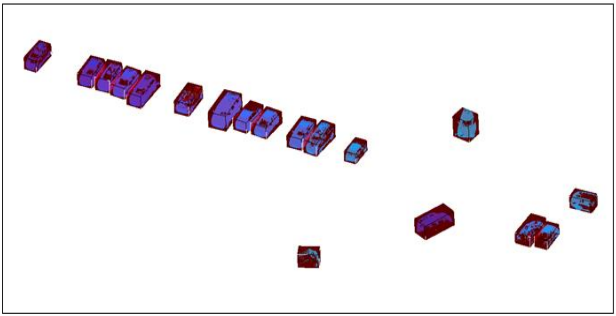

**Figure 11**. 3D bounding box around cars

The obtained bounding boxes from the reference point cloud are set as occupied parking spaces. Further, the classified Velodyne point cloud is used to examine whether the occupancy of defined parking spaces is valid or not at another point of time. For that purpose, a spatial analysis is performed to detect the Velodyne points labelled as cars inside the reference bounding boxes. If the Velodyne points are detected inside the bounding box, the parking space will be kept in the occupancy state, otherwise the space will be set as non-occupied. The velodyne scans will be also used for updating the reference map to always keep it up to date.

The 3D bounding box can then be used in the voxel environment of the initial map, to illustrate the occupancy of the parking spaces. Overall, the classification, clustering, and bounding box fitting process is used to extract useful information from point cloud data and visualize it in the voxel platform.

### 4.3 Visualization

For the visualization of the voxel-based urban environment the created 3D georeferenced grid of 10 cm dimension is transferred to the JavaScript library and application programming interface (API) "Three.js". The introduced API allows to create and display animated 3D computer graphics in a web browser with a further virtual reality extension possibility. To fit the rendering scene, the coordinates of the grid are additionally scaled and rotated for 90 degrees around X axis. Afterwards, the cube geometries of the corresponding dataset are generated using the Three.js library and placed according to the 3D grid corner positions. In this way, the urban environment is visualized by georeferenced voxels in a web browser. Furthermore, the derived bounding boxes are integrated in the rendered scene to represent the parking spaces. Figure 12 shows the voxel-based visualization with the parking spaces occupancy situation according to the reference data. In Figure 13, the changes in parking spaces' states after the revision according to the Velodyne data is visualized. There are three new free slots, which is indicated by the green boxes.

## 5. CONCLUSION

This paper has presented an approach to accurately estimate the position of a sensor platform using voxels. The voxelization process creates a regular 3D grid with 10 cm spacing, and a hierarchical approach is applied to down-sample the reference map as well as combined Velodyne scan to 1-meter voxels. The alignment of the combined point cloud with the reference point cloud is achieved using the RANSAC method. The resulting alignment is then refined using the point-to-plane ICP method, resulting in accurate alignment between the raw Velodyne scans and the reference map. The results showed that the estimation algorithm performs well overall, with most scans having relatively small errors. Furthermore, the paper discusses the potential application of the proposed method in the management of smart parking spaces, which is a major problem in big cities nowadays. This paper provides a basis for future studies on smart parking spaces and other urban application management. The approach recommended in the paper has the potential to be broadened to other applications and uses.

The following future work can be conducted to advance the scientific paper: scenes with less context information can be investigated to determine their effect on the accuracy of registration; scans obtained from different sensor position, such as UAVs, can be studied to evaluate the generalizability and robustness of the method; the influence of the size of the aggregated individual scans and hierarchy on accuracy and computational effort can be examined. Additionally, the labeling process can be explored to determine its impact on registration accuracy.

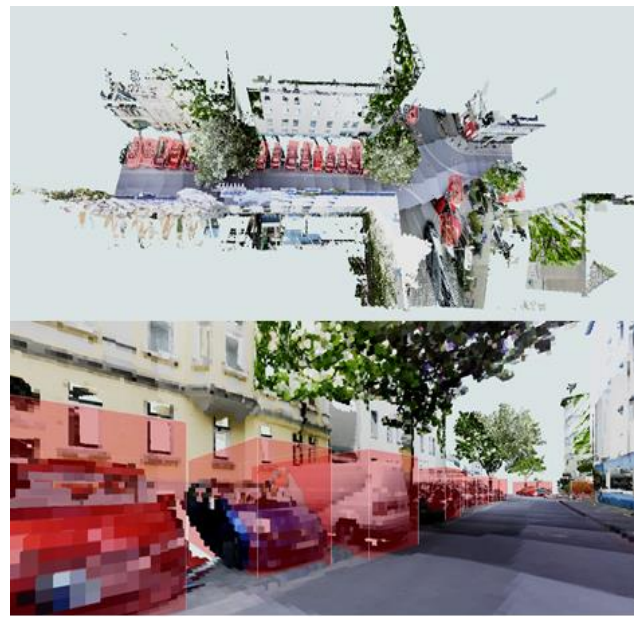

**Figure 12**. Voxel-based urban environment visualization with occupied parking spaces (red color)

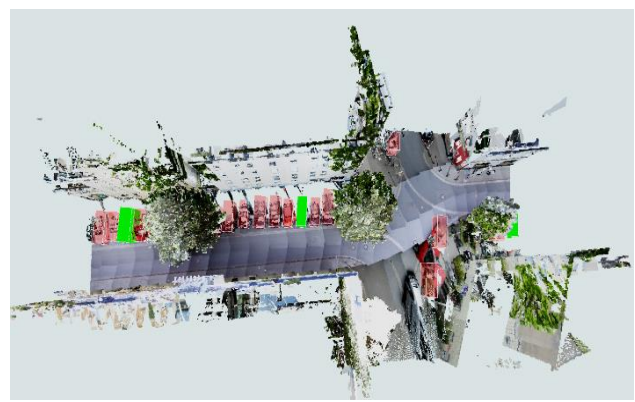

**Figure 13**. Voxel-based urban environment visualization with parking spaces states revised after the revision according to the Velodyne data (green color represents the vacant spaces)